## Cluster-based Specification Techniques in Dempster-Shafer Theory for an Evidential Intelligence Analysis of Multiple Target Tracks

Johan Schubert
*Division of Information System Technology, Department of Command and Control Warfare Technology, National Defence Research Establishment, S-172 90 Stockholm, Sweden. E-mail: schubert@sto.foa.se*

In Intelligence Analysis it is of vital importance to manage uncertainty. Intelligence data is almost always uncertain and incomplete, making it necessary to reason and taking decisions under uncertainty. One way to manage the uncertainty in Intelligence Analysis is Dempster-Shafer Theory [7]. This thesis [4, 5] contains five results regarding multiple target tracks and intelligence specification.

The objective of this work is to develop methods that will offer decision support to anti-submarine warfare intelligence analysts. The methods partitions intelligence reports into subsets. Each subset represents a possible submarine. It is able to reason simultaneously about the optimal number of submarines, which may be uncertain, and the optimal partition of intelligence reports among the submarines. Then, for each possible submarine a new algorithm calculates support and plausibility for possible tracks.

The first article [3] concerns the situation when we are reasoning with multiple events which should be handled independently. The idea is this. If we receive several pieces of evidence about different and separate events and the pieces of evidence are mixed up, we want to sort the them according to which event they are referring to. Thus, we partition the set of all pieces of evidence $\chi$ into subsets where each subset refers to a particular event. In Fig. 1 these subsets are denoted by $\chi_i$ and the conflict when all pieces of evidence in $\chi_i$ are combined by Dempster's rule is denoted by $c_i$. Here, thirteen pieces of evidence are partitioned into four subsets. When the number of subsets is uncertain there will also be a "domain conflict" $c_0$ which is a conflict between the current hypothesis about the number of subsets and our prior belief.

Now, if it is uncertain to which event some pieces of evidence is referring we have a problem. It could then be impossible to know directly if two different pieces of evidence are referring to the same event. We do not know if we should put them into the same subset or not.

To solve this problem, we can use the conflict in Dempster's rule when all pieces of evidence within a subset are combined, as an indication of whether these pieces of evidence belong together. The higher this conflict is, the less credible that they belong together. This was first suggested by Lowrance and Garvey [2].

Let us create an additional piece of evidence for each subset with the proposition that this is not an "adequate partition". Let the proposition take a value equal to the conflict of the combination within the subset. We call this evidence "metalevel evidence".

The article establishes a criterion function based on the "metalevel evidence". We will use the minimizing of this function as the method of partitioning the evidence into subsets. This method will also handle the situation when the number of events are uncertain.

The method of finding the best partitioning is based on an iterative minimization. In each step the consequence of transferring a piece of evidence from one subset to another is investigated.

After this, each subset of intelligence reports

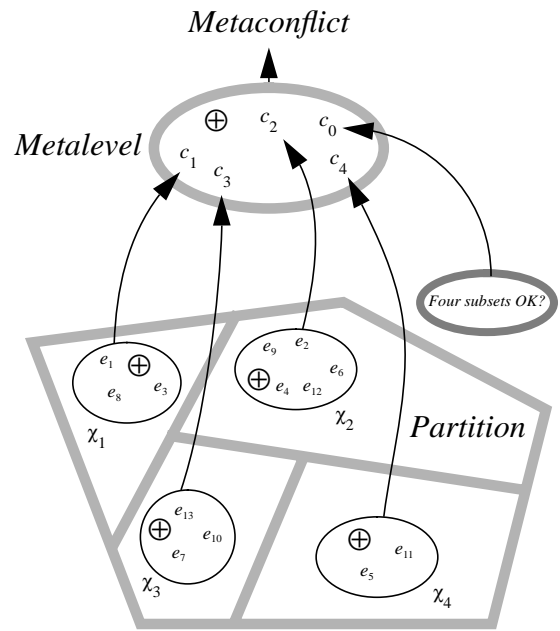

**Fig. 1.** The conflict in each subset of the partition becomes a piece of evidence at the metalevel.

refers to a different target and the reasoning can take place with each target treated separately.

In the second article we go one step further and specify evidence by observing changes in cluster and domain conflicts if we move a piece of evidence from one subset to another.

If some piece of evidence is taken out from a subset the conflict in that subset decreases. This decrease in conflict is interpreted as if there exists a pieces of metalevel evidence indicating that the piece of evidence that we took out does not belong to the subset where it was placed.

Similarly, if our piece of evidence after it is taken out from a subset is brought into any other subset, its conflict will increase. If a piece of evidence is moved in such a way that we receive a change in the number of subsets we will observe a change in domain conflict. These changes are interpret in a similar way.

We will find such evidence regarding each piece of evidence and for every subset.

When this has been done we can make a partial specification of each piece of evidence.

The extension in this article of the methodology developed in the first article imply that each piece of evidence will now be handled similarly by the subsequent reasoning process in different subsets if these subsets are approximately equally plausible. Without this extension the most plausible subset would take this piece of evidence as certainly belonging to the subset while the other subsets would never consider it at all in their reasoning processes.

In the third article [6] we set out to find a posterior probability distribution regarding the number of subsets.

We use the idea that each single piece of evidence in a subset supports the existence of that subset to the degree that it supports anything at all other than the entire frame.

We combine all pieces of evidence. From the result of that combination we can create a new bpa.

Where the previous bpa is concerned with the question of which subsets have support, the new bpa is concerned with the question of how many subsets are supported.

In order to obtain the sought-after posterior domain probability distribution we combine this newly created bpa with a given prior domain probability distribution.

The fourth article [1] derives a special case algorithm making it computationally feasible to analyze the possible tracks of a target. When it is uncertain whether or not the propositions of any two pieces of evidence are in logical conflict we may model this uncertainty by an additional piece of evidence against the simultaneous belief in both propositions and treat the two original propositions as nonconflicting. This will give rise to a complete directed acyclic graph with the original pieces of evidence on the vertices and the additional ones on the edges.

We may think of the vertices as positions in time and space and the edges as transitions between these positions. A proposition of a piece of evidence on an edge may, for example, tell us that the time difference between the two positions may be to small in relation to their distance.

We are interested in finding the most probable completely specified path through the graph, where transitions are possible only from lower to higher ranked vertices.

The algorithm reasons about the logical conditions of a completely specified path through the graph. It is hereby gaining significantly in time and space complexity compared to the step by step application of Dempster's rule.

To make decisions under uncertainty is somewhat complicated in Dempster-Shafer Theory because of the interval representation. In one approach the decision maker uses some additional information or subjective views. An article by Strat [8] is an example.

In this method an expected utility interval is constructed for each choice. If the interval of one choice is included in that of another, it is necessary to interpolate a discerning point in the intervals. This is done by a parameter $\rho$.

When we have several choices they may be preferred in different intervals of $\rho$.

The fifth article in this thesis is concerned with a situation where several different decision makers might sometimes be interested in having the highest expected utility among the decision makers rather than trying to maximize there own expected utility. For an individual decision maker we must here take into account not only the choices already done be other decision makers but also the rational choices we can assume to be made by later decision makers. The preference of each alternative to some decision maker is shown to be the length of the interval in $\rho$ where this alternative has the highest expected utility.

Department of Numerical Analysis and Computing Science, Royal Institute of Technology, Stockholm, Sweden.